\documentclass[letterpaper, 10 pt, conference]{IEEEtran}  % Comment this line out if you need a4paper

\IEEEoverridecommandlockouts                              % This command is only needed if
\title{\LARGE \bf
BiFNet: Bidirectional Fusion Network for Road Segmentation
\thanks{This work is supported by the Beijing Science and Technology Plan under Grants Z191100007419002,
and the National Natural Science Foundation of China (NSFC) under Grants No. 61803371.
}
}

\author{ Haoran Li, Yaran Chen, Qichao Zhang, Dongbin Zhao \emph{Fellow, IEEE}% <-this % stops a space
\thanks{Haoran Li, Yaran Chen, Qichao Zhang and Dongbin Zhao are with
the State Key Laboratory of Management and Control for Complex Systems,
Institute of Automation, Chinese Academy of Sciences, Beijing, 100190, China,
and also with he College of Artificial Intelligence, University of Chinese Academy of Sciences,
Beijing 100049, China. (email : lihaoran2015@ia.ac.cn, chenyaran2014@ia.ac.cn, zhangqichao2014@ia.ac.cn, dongbin.zhao@ia.ac.cn)}% <-this % stops a space
}%

\usepackage{amsmath,amssymb,amsfonts}
\usepackage{algorithmic}
\usepackage{graphicx}
\usepackage{amsmath}
\usepackage{textcomp}
\usepackage{threeparttable}
\usepackage{multirow}
\usepackage{booktabs}

\DeclareMathOperator{\atantwo}{atan2}

\begin{document}

\maketitle
\thispagestyle{empty}
\pagestyle{empty}

%%%%%%%%%%%%%%%%%%%%%%%%%%%%%%%%%%%%%%%%%%%%%%%%%%%%%%%%%%%%%%%%%%%%%%%%%%%%%%%%
\begin{abstract}
  
  % Road segmentation plays an important role in the intelligent driving system since
  % it provides a drivable area.
  % Due to the camera's sensitivity to illumination, image-based road detection methods
  % face enormous challenges in complex road conditions.
  % Since the light detection and ranging(LiDAR) data is not sensitive to illumination, road detection based on LiDAR and camera
  % fusion has gradually become the focus of research.
  Multi-sensor fusion-based road segmentation plays an important role in the intelligent driving system since
  it provides a drivable area.
  The existing mainstream fusion method is mainly to feature fusion in the image space domain
  which causes the perspective compression of the road and damages the performance of the distant road.
  % The main problem of these methods is that the perspective compression of the road in image space
  % causes the algorithm to focus more on the near road and ignore the distant road.
  Considering the bird's eye views(BEV) of the LiDAR remains the space structure in horizontal plane,
  this paper proposes a bidirectional fusion network(BiFNet) to fuse the image and BEV of the point cloud.
  The network consists of
  two modules: 1) Dense space transformation module, which solves the mutual conversion between camera image space and BEV space.
  2) Context-based feature fusion module, which fuses the different sensors information based on the scenes from
  corresponding features.
  % implements the scene context-based feature fusion.
  This method has achieved competitive results on KITTI dataset.
  % Furthermore, the fusion module is easily embedded in advanced 3D object detection methods.

\end{abstract}

\begin{IEEEkeywords}
  multi-sensor fusion, road segmentation, adaptive learning, autonomous vehicles
\end{IEEEkeywords}

\section{Introduction}
\IEEEPARstart{R}{}oad segmentation is a fundamental and essential task for an intelligent driving system, which provides
a driving area for self-driving system.
% With the help, the driving system locates itself in lanes, plans the feasible path and avoids the obstacles.
A robust road segmentation method is the precondition for driving system safety\cite{8119987}\cite{8303759}\cite{8962258}.
With the development of the deep neural networks,
deep neural network is also widely used in the perception\cite{7580631}, decision\cite{8789673}\cite{8852110} and control\cite{7592457}\cite{8360135} module of autonomous vehicles.
The road segmentation task based on deep convolutional
neural networks has been one of the research highlights for intelligent driving.

Currently, many works focus on camera image-based and LiDAR-based methods.
The camera captures the rich texture of the road while the LiDAR
measures the spatial structure of the environment by scanning laser beams.
Benefiting from the development of semantic segmentation methods\cite{li2018an}\cite{lu2019graph}\cite{8584494} based fully convolutional neural networks\cite{long2015fully},
camera image-based road segmentation has achieved a tremendous advance.
However, the camera is light-sensitive and the image is vulnerable to the interference of illumination.
 The camera image-based methods do not work well in overexposed or dark environments.
As an active light sensor, the point cloud of LiDAR is insensitive to illumination.
\cite{caltagirone2017fast}\cite{8467496} showed that LiDAR performed well on road segmentation task.
Unfortunately, the lack of texture and short valid detection distance are the defects of the point cloud.
% In addition,
%  since the point cloud is sparsity, the valid distance of the point cloud from the HDL-64E
%  to detection is about 60 meters.
Hence, the combination of two types of sensor — cameras, and LiDAR — has emerged as one of the most popular approaches for the road segmentation task.

The fusion-based road segmentation methods perform well by integrating the rich texture with the camera
and accurate altitude with the LiDAR.
These methods can be divided into two categories: post-fusion and feature fusion methods.
Post-fusion methods fuse the defected results from each sensor individually and obtain better road segmentation result, such as LC-CRF method\cite{gu2019road}.
Note that the feature fusion methods
% complete fusion during the features extraction,
obtain a merged feature by combining the features extracted from images and point clouds.
Due to the combination of the texture and the accurate altitude information,
the feature fusion-based methods like LidCamNet\cite{caltagirone2019lidar} and PLARD\cite{chen2019progressive} are more adaptive and robust, and attract many researcher's attention.
% Luca \emph{et al.}\cite{caltagirone2019lidar} projected the point cloud to the camera space and made use of convolutional neural networks to extract the features of the point cloud. They designed cross
%  fusion module to fuse the features from different sensor data.
%  Zhe Chen \emph{et al.}\cite{chen2019progressive} indicated that the projection of point cloud confused
%  the road with the non-road area. They proposed a LiDAR adaptation module which transformed the point cloud
%  to the altitude difference image. Then they used the feature space adaptation module to fuse the features
%  from the altitude difference image and camera image.

\begin{figure}
  \centering
  \includegraphics[width=0.40\textwidth]{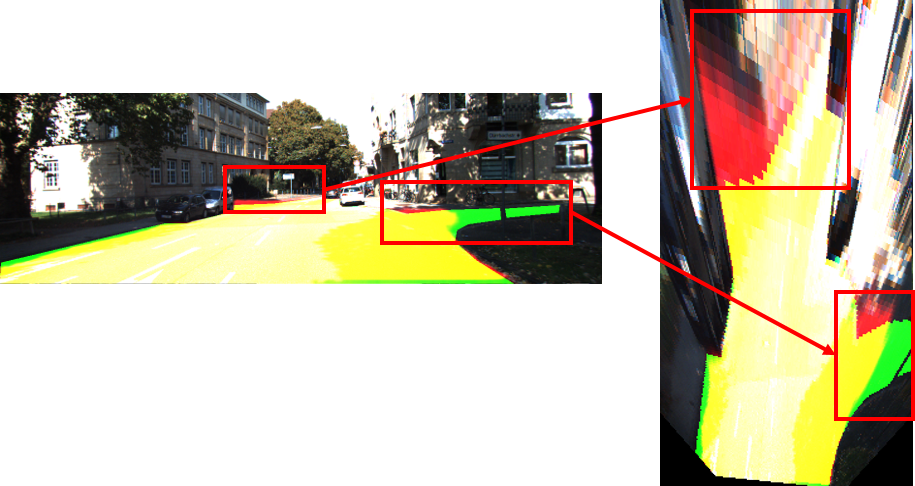}
  \caption{The influence of perspective compression to segmentation results.
  The left picture shows the segmentation result in the perspective space,
  and the right is the projected result in the BEV space. The yellow region is
  true positive(TP), the red region is false positive(FP), the green region is false negitive(FN).
  Since the farest objects are compressed sharply in perspective, adjacent pixels in the distance
  are segmented coarsely.}
\label{Fig_distance_compress}
\end{figure}

It should be mentioned that most feature fusion-based road segmentation methods complete feature fusion in camera space. According to the
 hole imaging theory, camera compresses the distant objects during the imaging processing which causes that the distant objects usually occupy few pixels than the near ones.
 Obviously, perspective compression increases the difficulty of the object segmentation in the distance.
 Fig. \ref{Fig_distance_compress} shows the imaging compression problem for road segmentation using the current feature fusion-based method.
 The road near can be segmented well, while the far one is segmented coarsely,
 due to the imaging compression.
%  which lead to the distribution
%  of the near objects and distant objects in camera space.
%  Therefore, the perspective projection based fusion detection method has better result on hand and
%  coarse result at the distance. The figure\ref{Fig_distance_compress} show the phenomenon.
As mentioned in \cite{caltagirone2019lidar}\cite{chen2019progressive}, how to combine the features which at same space is the important topic of feature fusion.
% Another important content for feature fusion is feature combination which adapts the different sensors
% to find more robust features\cite{caltagirone2019lidar}\cite{chen2019progressive}.

 Aiming at the above issues, we propose a road segmentation method which fuses the camera image and the BEV of the point cloud. On one hand, the BEV of the point cloud remains the distribution of road and 
 has enough information of segment road area. On the other hand, camera image has rich texture feature for road
 and far visible distance than LiDAR.
 Most methods focus on the feature fusion in camera space which use perspective projection and losses the spatial structure of
 the point clouds.
 In this paper, we design a dense space transformation module which transforms the features between the camera space
 and BEV space. Next, compared with the most current methods which fuse the features only based on
 position,  we propose a context-based fusion module which combines the features suitably and
fuses the transformed features adaptively according to the context.
To summarize, this paper presents the following contributions:
 \begin{itemize}
   \item We design a dense space transformation which builds the dense mapping between the image and BEV of the point cloud. This transformation is the foundation for fusing features from different spaces.
   \item We propose a context-based fusion module. This module fuses the multi-sensors features according to the environmental context and achieves the robust representation of the environment.
   \item Based on the above modules, we construct the bidirectional fusion network(BiFNet) which combines the camera image and BEV of the point cloud to implement road segmentation and achieves the competitive results in the KITTI road dataset.
 \end{itemize}

This paper is organized as follows. Section II describes
related works in road segmentation and the sensor fusion.
Section III presents our definition and analysis about the sensor fusion task.
Then, we give the details of our proposed fusion module includes dense space transformation and
attention based feature fusion module in Section IV.
Furthermore, the ablation study and the comparison experiment are displayed and
the analysis of methods is given. Finally, in
Section V, we summarize our work and discuss future work.

\section{Related Work}

\subsection{Image-based road segmentation}
Current image-based methods can be divided into two categories: model-based and learning-based.
Model-based methods\cite{8851512}\cite{4621283}\cite{1706865} build the road model based on unique shapes and textures.
Mohamed \emph{et al.}\cite{aly2008real} detected the lanes according to the shape color and figured out the road area in the image.
Ankit \emph{et al.}\cite{laddha2016map} used the appearance features of the road surface and map prior to
implement the automatic labeling road area.
The model built by expert domain knowledge needed to tune according to the different datasets or scenes.
Model-based methods rely on the robust road model which are applicable to scenes with the similar road.
For the complex illumination and textures, these methods have bad adaptive ability.

Learning-based methods are popular at road segmentation task. These methods learn the robust classifier
to identify the road pixel in the image fed by large labeled datasets. Profiting by the development of
deep learning and convolutional neural networks, hand-designed features\cite{zhou2010road} used for classifier are replaced
with automatic learning features\cite{long2015fully}\cite{oliveira2016efficient}.
At the same time, the development of semantic segmentation methods has greatly advanced road detection methods.
Since the road has a special appearance, Zhe \emph{et al.}\cite{chen2017rbnet} added the road boundary constraints to
the deep segmentation networks and improved the performance of methods.

\subsection{LiDAR-based road segmentation}
Although image-based segmentation methods have developed rapidly in recent years, the image noise caused by
 the illumination leads to the limitation of the image-based methods. As a distance sensor, LiDAR is not
 affected by light, and many researchers use the point cloud to detect road.
 Fernandes \emph{et al.}\cite{fernandes2014road} introduced the similarity of the histogram of the point cloud
 as the discriminator to classify the points and segmented the road from the point cloud.
 Caltagirone \emph{et al.}\cite{caltagirone2017fast} employed the fully convolutional neural networks over
 the BEV of the point cloud and achieved the desired result.
 Zhang \emph{et al.}\cite{8291612} proposed a sliding-beam method to segment the road by using the off-road data and applied
 a curb-detection method to obtain the position of curbs for each road segments.

 \subsection{Camera and LiDAR fusion-based road segmentation}
%  Although LiDAR has the advantage of being insensitive to lighting conditions,
%  its disadvantages are also very obvious. The point cloud of LiDAR is sparse and the features are
%  textureless which include only the spatial position and reflection intensity information of the point.
%  Since camera image contains rich texture features,
Depth information is the key to solve the problem of RGB image.
How to combine depth information with color image effectively has always been a research hotspot\cite{8026575}\cite{8091125}.
 Since the data of the camera and LiDAR can complement each other,
 fusing image and point cloud to segment road
 has gradually become the mainstream  method. Shinzato \emph{et al.}\cite{shinzato2014road} proposed a fusion method
 which combined the distribution of the point cloud and image pixels as the discriminator. Conditional
 Random Field(CRF) was used for constructing the fusion model. There are many methods\cite{xiao2015crf}\cite{han2017road}\cite{xiao2018hybrid} which constructed
 Gibbs energy by using Euclidean distance between the points and RGB values of pixels.
 The above methods needed the expert knowledge to build the data features and combined the features
 to find robust classifier for the pixels.
 Gu \emph{et al.}\cite{gu2019road}\cite{8370690} used the CRF to fuse
 the results from the point cloud\cite{8525305} with image features extracted by deep convolutional neural networks.
 Compared to the hand-selected features, this method achieved better performance.
 Caltagirone \emph{et al.}\cite{caltagirone2017fast} projected the point cloud into camera space and
 designed a cross fusion module to fuse image features with front view features of the point cloud.
 Since the coordinate of the point can not directly reflect the road characteristics, Chen \emph{et al.}\cite{chen2019progressive}
 proposed an altitude difference transformation which transformed the point cloud to a special front view image.
 And they introduced the feature space adaptation module to fuse the features from different sensors.

 \subsection{Camera and LiDAR fusion-based 3D object detection}
 Image-based 2D object detection and tracking have made great progress\cite{ren2015faster}\cite{liu2016ssd}\cite{redmon2017yolo9000}\cite{chen2018temporal}.
 Since the point cloud of LiDAR has accurate distance information\cite{li2019comparison}, fusing the camera image with the point cloud also
 plays an important role in 3D object detection. The categories of the fusion for object detection
 can be divided into a region-based method and point-based method.
 The region-based method combined the sensors data or features from a unique 3D region and got a better
 feature representation\cite{chen2017multi}\cite{ku2018joint}.
%  Chen \emph{et al.}\cite{chen2017multi} proposed a region-based fusion network
%  which fused the multiple views from different sensors. This method is widely used in 3D object detection\cite{ku2018joint}.
 Compared with the region-based methods, point-based methods fused the features in terms of the point and the pixel.
 Continuous fusion layer\cite{liang2018deep} belongs to this category.
 It built the relationship of pixels in the camera image and the BEV-based on 3D points and
 designed the continuous fusion which constructed a mapping between sparse points and BEV pixels.
 You \emph{et al.}\cite{you2018pvnet} proposed an attention fusion block which cascaded the global context feature of
 camera image and feature of each point. PointFusion\cite{xu2018pointfusion} used such a method to implement 3D object detection.
 Wang \emph{et al.}\cite{wang2018fusing} proposed a sparse non-homogeneous pooling and used projection matrix
 to transform features between the RGB image and the BEV.

 \section{Problem Formulation}
 Road segmentation refers to the analysis of sensor data by an algorithm to detect the drivable road area
 in the current environment. Supposing that $D$ is the sensor data, $G$ is the corresponding ground truth.
%  The objective of segmentation is finding the predictive model $\phi$ which minimizes the segmentation loss
%  under current sensor data distribution.
%  \begin{equation}
%   \min_W  \mathcal{L}(\phi(D; W), L).
% \end{equation}
% Here $W$ is the parameter of the predictive model and $\mathcal{L}(\cdot, \cdot)$ is the loss function.
% Because of the limitation of each sensor, multi-sensor fusion based road segmentation fuses individual sensor
% data to obtain better features and achieves more robust segmentation results.
Due to the difference in sensor measurement accuracy and imaging model, the key to multi-sensor
fusion is to find the optimal combination between sensors to complete the segmentation task.
% With all sensor data in the current environment, the objective of multi-sensor based road segmentation
% is find out the predictive model $\phi$ which minimize the loss function.
% \begin{gather}
%   \min_W  \mathcal{L}(\phi(D^1, D^2, \cdots, D^s; W), L)
% \end{gather}
% Here $D^s$ is $s$-th sensor data.
For feature-based fusion methods, we need to determine two models: classification model $\phi _c$
and fusion model $\phi _f$, by minimizing the objective
\begin{gather}
  \min_{W, \Theta}  \mathcal{L}(\phi_c(\phi_f(D^1, D^2, \cdots, D^s; \Theta), W), G),
\end{gather}
where $\Theta$ and $W$ are weights of fusion model and classification model, respectively.
Since single sensing has its limitations, such as camera image is sensitive to light and
the point cloud of LiDAR is sparse and textureless. It is difficult to cope with complex scenes
with a single sensor. The purpose of multi-sensor fusion is to combine the advantages of different
sensors to make up for the disadvantages of a single sensor and achieve complementary advantages.

Recently, many works focus on identifying the fusion model $\phi _f$ which can be divided into
three categories including pre-fusion, post-fusion, and feature fusion.
The key of pre-fusion is converting different sensor data into a unified data format. Then it uses
a single model to process the cascaded data. The advantage of this method is that it only needs
to preprocess the data, no special design of the detection model is needed. The shortcoming is obvious.
During the data preprocessing, the processed data losses the characteristics of the original data space.
For example, the point cloud makes a perspective projection to the camera space, and the projected
data structure cannot maintain the geometric spatial distribution of the point cloud data.

For post-fusion, the fusion model is designed for the detection results of different sensors to obtain
the final fusion result. In this case, each sensor requires a separate design detection model.
This method is often used in object detection and segmentation. For object detection, the most
common post-fusion method is Bayesian filter. For the segmentation task, the common is CRF.

Feature fusion combines different sensor data features to form a better scene representation,
and the combined features are used for road segmentation.
% Feature extraction is done on the respective raw data space.
% This method is developed in recent years and is widely used in 3D object detection.
% Compared with object detection,
The segmentation task has higher precision requirements for feature fusion
and needs to achieve alignment of points in different data spaces to achieve point-wise feature fusion.

% ====================================== this is for figure  =========================================
\begin{figure*}
  \centering
  \includegraphics[width=0.98\textwidth]{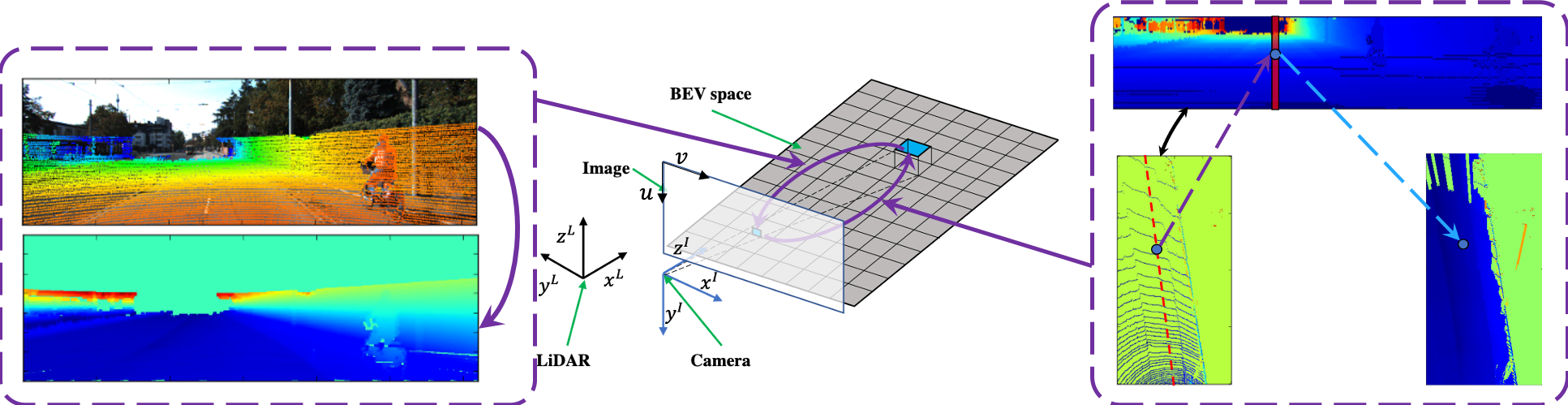}
  \caption{The dense upsampling during the transformation between the perspective space and the BEV space.
  The left part shows the dense sampling under perspective space,
  which is needed to establish the index relationship from perspective space to BEV space.
  The right part shows the dense sampling process of the height under the BEV space,
  which is needed to establish the index relationship from the BEV space to the perspective space.
  The middle part represents the 3D point to camera projection process  and the relationship between camera and LiDAR coordinate system.
  The blue arrows mean the camera coordinate where the directions of $x^I$, $y^I$ and $z^I$ are left, down and forward, respectively.
  The black arrows represent the LiDAR coordinate where the directions of $x^L$, $y^L$ and $z^L$ are forward, right and up, respectively.
  % The camera captures the object in 3D space and generates image by pinhole imaging.
  % The LiDAR measures the 3D spatial position directly. The dark gray plane means the
  % BEV space. And the purple curve is the transformation between the perspective space and BEV space.
  }
\label{Fig_space_transform}
\end{figure*}
% ========================================= end for figure ==========================================

In this paper, we propose the BiFNet for road segmentation and mainly solve the two problems of multi-sensor feature fusion
\begin{itemize}
  \item Aligning the features from different data space. Because of the different working principles of
  different sensors, the content, dimension and the data structure of the different sensors data are various.
  % For example, image data focuses on the texture representation of the environment,
  % while point cloud data focuses on the geometric structure of the environment.
  The premise of feature fusion is that feature expression needs to be in the same data space and feature space.
  Therefore, feature alignment in data space is the base of feature fusion.

  \item Finding out the optimal combination of features. Different sensors have different expressive abilities
  for different scenes. For example, the expressive abilities of images for high exposure
  or very dark scenes are relatively weak, while point cloud is not affected by illumination factors.
  % On the other hand, the point cloud can not distinguish road areas according to texture features,
  % such as traffic signs (lane lines) and achieve road segmentation.
  % Therefore, in the scenario where only traffic signs are used to distinguish road areas,
  % the expression ability of point cloud is weak.
  The purpose of feature combination is to find a combination that can adaptively combine
  the features of different sensors to form the most robust feature according to
  the different scenes.
\end{itemize}

To facilitate the understanding of the algorithm, we list the key parameters symbols in Table \ref{table_para_alg}.

\begin{table}
  \centering
  \caption{Key parameters symbols of Bidirectional Fusion Network.}
  \scalebox{1.0}{
  \begin{tabular}{c c}
      % \hline
      \toprule[1.5pt]
      Symbol    & Implication  \\
      \hline
      $D$   & Sensor data  \\
      % \hline
      $\Theta$,$W$ & Weights of the fusion and classification models \\
      % \hline
      $G$ & Ground truth  \\
      % \hline
      \multirow{2}{*}{$x^L$, $y^L$, $z^L$} & Axises of LiDAR\\
                                  & Superscript $L$ means LiDAR coordinate \\
      % \hline
      \multirow{2}{*}{$x^I$, $y^I$, $z^I$} & Axises of Camera  \\
                                  & Superscript $I$ means camera coordinate\\
      % \hline
      $(u,v)$ & Pixel in image  \\
      % \hline
      $\Delta \theta$, $\Delta \phi$ & LiDAR angular resolutions \\
      % \hline
      $R$, $T$ & External parameters \\
      % \hline
       $K$ & Intrinsic matrix \\
      % \hline
      $\bar{U}$ & Homogeneous camra coordinates \\
      % \hline
      $\bar{\Gamma}^L$ & Homogeneous 3D coordinates in LiDAR \\
      % \hline
      $\hat{\Gamma}^L$ & Pseudo 3D coordinates in LiDAR \\
      % \hline
      ${\Gamma}^L$ & 3D coordinates in LiDAR \\
      % \hline
      $f$ & Feature map \\
      % \hline
      $\kappa$ & Feature representation ability \\
      % \hline
      $\hat{\mu}$ & Context quality \\
      % \hline
      \multirow{2}{*}{$\Theta^{CBF}$, $\Theta^{DT}$} & Weights of context-based fusion \\
                                                     & and domain transformation \\
      % \hline
      $\xi(\cdot)$ & Distance function \\
      % \hline
      $\sigma(\cdot)$ & Sigmoid function \\
      % \hline
      $\psi(\cdot)$ & Multilayer perceptron \\
      \bottomrule[1.5pt]
  \end{tabular}
  }
  \label{table_para_alg}
\end{table}

\section{Bidirectional Fusion Network}
For feature alignment problem, we propose a dense space transformation module which transforms
the features between camera space and BEV space. At the same time, we design a feature fusion
module based on the channel attention mechanism, which fuses the transformed features adaptively
according to the scene context.
Based on these two modules, we construct the bidirectional fusion network which realizes the
end-to-end feature fusion road segmentation based on the camera image and the point cloud.
% The method extracts features from the camera space and BEV space of the point cloud separately
% and fuses the features through the bidirectional feature fusion module, thereby realizing
% road segmentation in a complex environment.

\subsection{Dense space transformation(DST)}
The image and point cloud can be obtained by observing the objects in 3D space with different sensors.
Image is the 2D data of the camera through the principle of the hole imaging, compressing
the spatial structure and retaining the texture.
% -------
Far objects will scale proportionally on the imaging plane.
The far road will be compressed and the proportion is much smaller than the near road proportion,
which results in the perspective image
based methods paying more attention to the accuracy of the near road.
% -----
Point cloud is the spatial data obtained by LiDAR through scanning a beam of laser to image
an object. It retains the spatial structure but loses texture.
% ---
% Point cloud is unordered data\cite{qi2017pointnet}. The order of the points does not affect the attributes. 
Chen \emph{et al.}\cite{chen2017multi} proposed two projection methods which converted
the point cloud into ordered data which we use to upsample the height.
% ---
BEV is one of the common preprocessing methods used in road segmentation.
It compresses only the height of the point cloud and preserves the geometry of the road.
In this section, we build the relationship between the camera space and BEV space as shown in Fig. \ref{Fig_space_transform}.
%  and transform the features between these spaces as shown in Fig. \ref{Fig_space_transform}

\subsubsection{Upsample perspective height}
In order to achieve dense mapping from BEV space to camera space, the height of
each pixel in camera image is needed. Then according to the pinhole camera model, we calculate the 3D
coordinate of each pixel with the 3D point cloud. However, the point cloud and perspective projection are sparse. In order
to obtain the dense height map, we upsample the sparse perspective projection as shown in left part of Fig. \ref{Fig_space_transform}.
$(u, v)$ is the position of the pixel in the image. The valid projection neighbourhood region of the image pixel
is $\mathcal{N}$, and the estimated height $\hat{z}^L$ of this pixel in LiDAR coordinate is
\begin{equation}
  \hat{z}^L = \frac{1}{\Xi} \sum_{(u_n,v_n) \in \mathcal{N}}  \frac{z^L_n}{\sqrt{(u_n - u)^2 + (v_n - v)^2}}.
\end{equation}
Here $\Xi$ is the parameter for normalization. $u_n, v_n$ are the neighbour positions in the image,
$z^L_n$ is the height of the neighbour in 3D space.

\subsubsection{Upsample BEV height}
In order to achieve dense mapping from camera space to BEV space, It is required to know the spatial position
of each pixel in BEV. Due to the sparsity of the point cloud, only a small portion of the pixels
in BEV has a valid height value. In order to obtain the dense height of BEV, according to the principle of
rotating scanning laser imaging, a height sampling method in BEV space is designed.
Firstly, all points are ranked in the point cloud according to \cite{chen2017multi} and calculate
the index values $r, c$ of each point.
% --
\begin{align*}
  c &= \lfloor \atantwo (y^L, x^L) / \Delta \theta \rfloor \\
  r &= \lfloor \atantwo (z^L, \sqrt{{x^L}^2+{y^L}^2}) / \Delta \phi \rfloor
\end{align*}
Here $x^L, y^L, z^L$ are the coordinates of each point. $\Delta \theta$ and $\Delta \phi$ are
the horizontal and vertical angular resolutions of the LiDAR, respectively. % According to this formula,
% --
For each pixel on the BEV, we calculate the corresponding column $c$ in arrangement map $M$ which is corresponding to the top picture in the left part of Fig. \ref{Fig_space_transform}.
$M_{c}$ is the $c$-th column from the arrangement map, corresponding to the red column
in the right part of Fig. \ref{Fig_space_transform}. It is required to find out the neighbour $n_i$ of the
BEV pixel in $M_{c}$.
\begin{equation}
  \xi (n_i)>= \xi (p_i), \text{ } \xi (n_i+1)<= \xi (p_i)
\end{equation}
Here $\xi(\cdot)$ is the distance between the projected point and LiDAR. $p_i$ is the pixel in BEV whose height is needed to estimate.
And the estimated height of this pixel is
\begin{equation}
  \hat{z}^L_{p_i} = \frac{\xi(n_i, p_i)}{\xi(n_i, n_i+1)} z^L_{n_i} + \frac{\xi(n_i+1, p_i)}{\xi(n_i, n_i+1)} z^L_{n_i+1}.
\end{equation}
Here $\xi(\cdot, \cdot)$ is horizontal distance between two points. $z^L_{n_i}$ is the valid height in arrangement map,
and $\hat{z}^L_{p_i}$ is the estimated height.

\subsubsection{Camera to BEV space transformation}
Giving the features in camera space, the goal of the camera to BEV space transformation is
to calculate the corresponding features in BEV space. The rotation matrix of the LiDAR relative
to the camera is $R$. The transformation matrix is $T$. The intrinsic matrix of the camera is $K$.
According to the pinhole camera model, we can calculate the pseudo spatial position $\hat{\Gamma}^L$ of the pixel
by inverse transform.
\begin{equation}
  \hat{\Gamma}^L = R^{-1} K^{-1} \bar{U}.
\end{equation}
Here $\bar{U}$ is the homogeneous coordinates of the pixel in the camera image. Since camera imaging is nonlinear
transformation, there is a scale between the pseudo spatial position and the real spatial position.
We can estimate the scale $\tau$ by
\begin{equation}
  \tau = \frac{\hat{z}^L + R^{-1} T |_z}{z^L},
\end{equation}
where $\hat{z}^L$ is the $z$ axis value of $\hat{\Gamma}^L$. And $R^{-1} T |_z$ means the $z$ axis value of $R^{-1} T$.
And the final 3D spatial position $\Gamma^L$ is
\begin{equation}
  \Gamma^L = \tau (\hat{\Gamma}^L + R^{-1} T).
\end{equation}

% --
% When the feature map $f$ is obtained by downsampling $s$ times the image, each pixel in $f$ is
% corresponding to a region in original image. The coordinates of the upper left corner and the
% lower right corner of the region are $[sx, sy]$ and $[(s+1)x, (s+1)y]$, respectively.
% The region $M_b$ in BEV is $[z_t/((s+1)r), y_t/((s+1)r)] - [z_t/sr,y_t/sr]$. We employ
% average pooling to convert the region features to the pixel feature.
% \begin{equation}
%   f_i = \frac{1}{M_p} \sum_{M_t} f_t
% \end{equation}
% Here $M_p$ is the parameter for normalization. $f_t$ and $f_i$ are the source feature and target feature,
% respectively.
% --

\subsubsection{BEV to camera space transformation}
In order to complete the feature transformation from BEV to camera space, we only need to estimate the
spatial position of each pixel on BEV map. According to the procedure of constructing BEV,
The horizontal coordinates $x^L, y^L$ of each pixel in the 3D space are linearly related to the coordinates in BEV.
Moreover, the dense BEV height map have been calculated in previous works. Then, we build the mapping
from BEV space to camera space by hole imaging theory.
\begin{equation}
  U = K [R \text{ } T] \bar{\Gamma}^L
\end{equation}
Here $\bar{\Gamma}^L$ is the homogeneous coordinates of the pixel in the 3D space.

\subsection{Context based fusion(CBF)}
\begin{figure}
  \centering
  \includegraphics[width=0.45\textwidth]{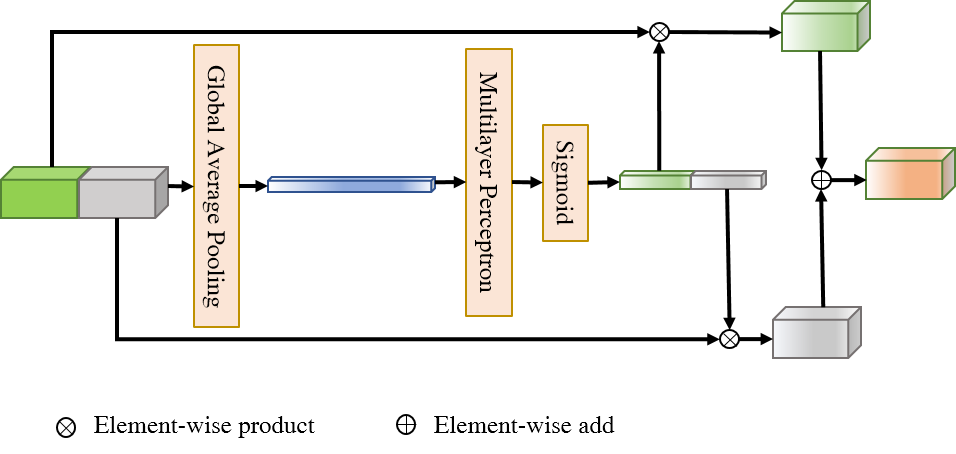}
  \caption{The context-based fusion module(CBF). The left green cuboid means the feature map from image,
  and the gray means LiDAR feature map. The right orange means the fused feature map.}
\label{Fig_fusion_module}
\end{figure}

Since the imaging principle and measuring range of each sensor are different, the abilities of sensors to
express various environments in diverse scenarios are different.
% For dark or overexposed scenes, the ability of image expression is weak.
% For scenes where the geometrical structure of the environment is not obvious, the expression ability of point clouds is weak.
The purpose of the multi-sensor fusion is obtaining the consistent and effective
representation of the environment for various scenarios.
In order to achieve this effect, the direct methods are projecting the LiDAR features into camera space and
adding the features into image features.
% LidCamNet adds the features of different sensors directly,
% in which the weights and translation parameters are used to adapt to the different distributions
% of the features of different sensors.
% Compared with LidCamNet, PLARD predicts scale and translation parameters related to features of
% image and point cloud  for each feature vector in the LiDAR feature map,
% so as to realize the adaptive fusion of feature at the pixel level.
The problem of this direct linear addition method is that it does not consider the different expressive abilities of
 sensors for various scenarios, which may result in the confusion of features when the difference
between the two features is large.
Inspired by SENet\cite{hu2018squeeze}, we design a context-based adaptive feature fusion module,
as shown in Fig. \ref{Fig_fusion_module}.
According to the scene expression ability of different sensor features,
the module can stimulate the good features and suppress the bad features,
so as to get a more effective feature expression.

% First, we need to evaluate the ability of different features to represent scenes.
For the feature maps from convolutional neural networks, we argue that
different channel features represent different context of the environment.
% and all features are connected in series to form the feature representation of the current environment.
% We calculate the expressive ability of the current feature channel by global pooling.
We can get the feature representation ability $\kappa$ of the current feature map as follows
\begin{equation}
  \kappa = \frac{1}{H \times W} \sum^{H}_{h=1} \sum^{W}_{w=1} f_{h,w,*}.
\end{equation}
Here $H$ and $W$ are the height and width of the feature maps, respectively.
$f_{h,w,*}$ is the corresponding vector at $(w, h)$ of the feature map.
% $\kappa$ is a statistic of the current characteristic channel $c$,
which is insensitive to noise compared with \emph{max pooling}.

After obtaining the expressive power of different feature channels, we need to determine which channels are conducive to the segmentation task
and which ones confuse features. SENet pointed out that suppressing the poor features and
encouraging the better feature is conducive to more effective feature expression.
Therefore, we need to consider the relationship between channels,
to evaluate the favorable feature channels and the disadvantageous feature channels.
Here we use a three-layer perceptron(MLP) to predict the gain of each channel for
the current segmentation task in the current scenario.
\begin{equation}
  \hat{\mu} = \sigma( \psi ( \psi(\kappa; \Theta^{CBF}_1); \Theta^{CBF}_2)),
\end{equation}
where $\hat{\mu}$ represents context quality for the features.
And the $\sigma(\cdot)$ sigmoid function at the output layer is used to scale the weights into $[0, 1]$.
$\kappa$  is the feature representation ability vector.
$\psi$ is the fully connected layer, and $\Theta$ means the weight of the layer.
At the middle layer of MLP, we use ReLU as the activation function.
And the final feature in each channel is
\begin{equation}
  \hat{f}_c = \hat{\mu}_c \times f_c.
\end{equation}
Here $f_c$, $\hat{\mu}_c$ and $\hat{f}_c$ are the original features, context quality and reweighted features of the channel $c$, respectively.
For the feature of image and point cloud, we fuse the weighted features based on context
 to get the final consistent feature.
%  \begin{equation}
%   f^F = \hat{f} ^I + \hat{f} ^L.
% \end{equation}

\begin{figure*}
  \centering
  \includegraphics[width=0.9\textwidth]{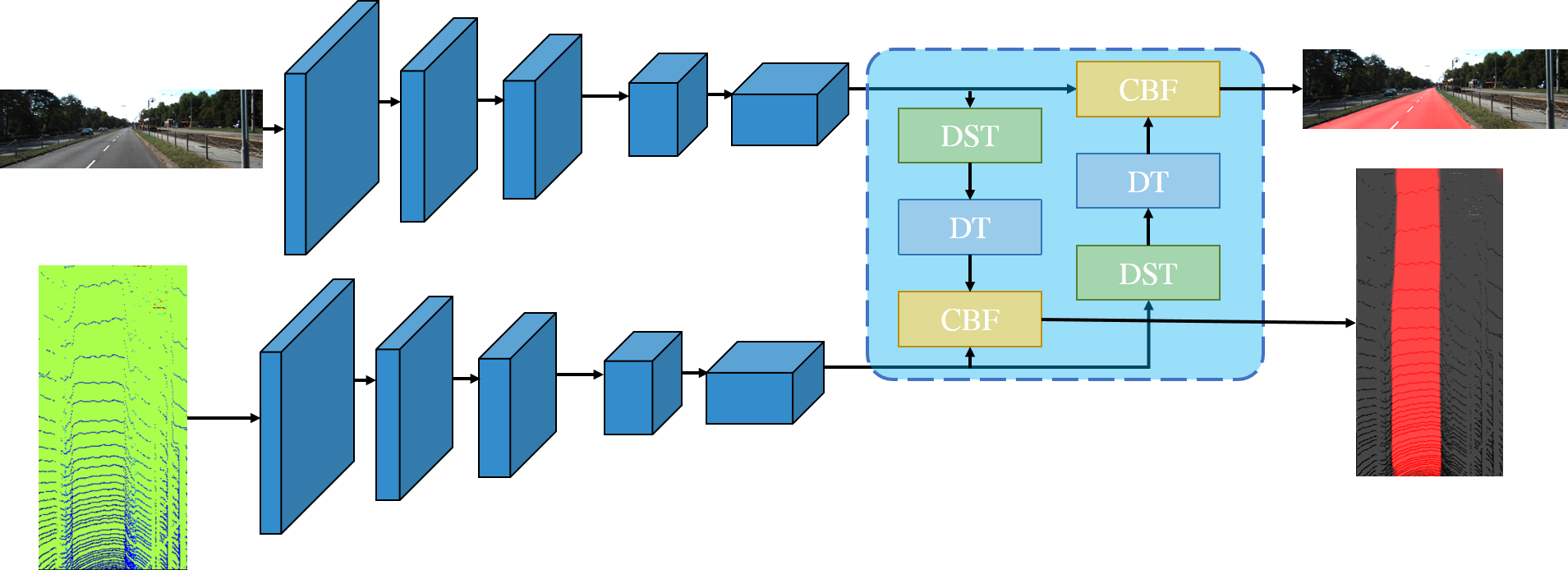}
  \caption{The BiFNet architecture.
  The networks have two backbones. Each backbone has five convolution blocks same as ResNet.
  The module has two groups, and each group is composed of DST, DT and CBF.
  DST module transforms the features into the same spatial space, and DT adapts the appropriate
  domain since the multi-sensors have different measurements and different feature dimensions.
  Finally, CBF fuses the features which have the same spatial space and domain. The output of the camera space branch
  is the final result for testing.}
\label{Fig_network}
\end{figure*}

\subsection{Bidirectional fusion networks}
Based on the dense space transformation module and the context-based fusion module, we design the
bidirectional fusion module to fuse the features from different deep features.
Considering the different imaging modes of images and point clouds,
their data sources and data expressions are also different.
There may be differences in scale and dimension arrangement between the two group features.
Therefore, we add a domain transformation(DT) module between the DST and CBF. And it is realized by $1 \times 1$
convolution.
\begin{equation}
  % X^D_{i, j} = X_{i, j} W
  f^D_{i, j, k} = \sum_c f_{i, j, c} \Theta^{DT}_{k, c}.
\end{equation}
Here $f_{i, j, c}$ is $i$-th row, $j$-th column and $c$-th channel element of the feature map.
$f^D_{i, j, k}$ is the corresponding domain transformed feature with $k$ channels.
$\Theta^{DT}$ is the parameter of the convolution.

Based on this module, we construct the bidirectional fusion networks. Since ResNet\cite{7780459} achieves
the excellent performance in segmentation tasks, we also use ResNet as the backbone
network to extract features of image and BEV of point cloud respectively.
After the output of two parallel ResNet, we insert a bidirectional fusion module to realize
feature fusion. The whole network architecture is shown in Fig. \ref{Fig_network}.

In order to make the network converge faster and better, we adopt the multi-task learning to train the network.
We introduce the road segmentation loss in camera space and BEV.
Note that the loss of camera space can make the network see objects farther away, while the loss of BEV
can make the distant and near objects distribute equally.
% \begin{align*}
%   \mathcal{L} =& - \frac{1}{B \times W_I \times H_I} \sum^B_{b=1} \sum^{W_I}_{m=1} \sum^{H_I}_{n=1} \hat{y}^I_{b,m,n} \\
%                & - \frac{1}{B \times W_L \times H_L} \sum^B_{b=1} \sum^{W_L}_{m=1} \sum^{H_L}_{n=1} \hat{y}^L_{b,m,n}
% \end{align*}
For each pixel in camera space, we use the focal loss\cite{Lin2017Focal} for the imbalance between road and background
\begin{align*}
  \mathcal{L}^I_p =&  \alpha^I (1 - \hat{y}^I_{p})^{\gamma} g^I_{p} \log(\hat{y}^I_{p}) + (\hat{y}^I_{p})^{\gamma}(1-g^I_{p}) \log (1-\hat{y}^I_{p}).
\end{align*}
Here $\mathcal{L}^I_p$ means the loss of the pixel $p$ in image $I$.
Where $\alpha$ reduces the impact of the imbalance between the classes. $\gamma$ controls the weight of the hard examples. $g$ is the pixel label,
and $\hat{y}$ is the estimated probability belonging to the road.
Multi-task learning are widely used for objects detection and segmentation\cite{chen2018multi}.
And we use the multi-task objective function to train the whole networks
\begin{equation}
  \mathcal{L} = \beta \sum^{N^I}_{p=1} \mathcal{L}^I_p + \sum^{N^L}_{p=1} \mathcal{L}^L_p.
\end{equation}
Here $N$ is the number of the  pixels in one batch. $\beta$ is the weight for balancing the different loss from
different space. The superscripts $I, L$ is corresponding to the camera space and LiDAR BEV space.

\section{Experiments}
In this section, we evaluate the method on KITTI\cite{Geiger2013IJRR} dataset.
Firstly, we briefly introduce the KITTI road segmentation data set and evaluation metrics.
The training process of the networks is given in detail.
Then, we evaluate the impact of each module proposed on the final results on the validation set
of the KITTI dataset and analyze the results.
Finally, we evaluate the algorithm on the test set of the data set.

\subsection{Dataset and evaluation metrics}
% KITTI is a very complete open data set of automatic driving scenes, including images, LiDAR,
% GPS and other data sets, and provides different data annotations for different tasks.
KITTI is a well-know dataset which provides 289 training images  and 290 testing images for road segmentation tasks.
The training set and the test set contain three different road scenarios including
Urban Marked road(UM), Urban Multiple Marked road(UMM), and Urban Unmarked road(UU).
% In the original data set, the size of each picture is slightly different.
% We reset the size of all the pictures to $384 \times 1248$.

In the KITTI dataset, the LiDAR data format is 3D point cloud.
Because of the disorder of 3D point cloud, convolutional neural networks can not be directly applied to point cloud.
As mentioned earlier, the LiDAR input part of our algorithm is the top view projection of 3D point cloud.
The resolution of the BEV is 0.5 meter. Each pixel in the overhead projection contains a vector,
which has 6 attributes, namely, the maximum height value, the minimum height value,
the average height value, the maximum intensity value,
the minimum intensity value and the average intensity value of the point set falling into the pixel area.
The default value is 0.

We use the evaluation metrics in literature\cite{fritsch2013new} to analyze our algorithm.
The evaluation measures include precision(PRE), recall(REC) and pixelwise maximum F-measure (MaxF).
PRE measures the proportion of correctly segmented pixels,
and REC reflects the missing detection of the algorithm.
The calculations are as follows.
\begin{align*}
  \text{PRE} &= \frac{TP}{TP+FP} \\
  \text{REC} &= \frac{TP}{TP+FN}
\end{align*}
Here $\text{TP}$, $\text{FP}$ and $\text{FN}$ are true positive, false positive and false negitive, respectively.
There is a bit of trade-off between PRE and REC. MaxF is designed as a comprehensive metric considering PRE and REC.
Another comprehensive metric is the average precision(AP), which provides insights into the performance over the full recall range.
The calculations are as follow.
% The evaluation measures used in the following comparisons are the pixelwise maximum F-measure (MaxF),
% precision(PRE), recall(REC), and average precision(AP). These metrics are from \cite{fritsch2013new}.
\begin{align*}
  \text{MaxF} &= \arg\max_{\tau} \frac{\text{PRE} \times \text{REC}}{\text{PRE} + \text{REC}} \\
  \text{AP} &= \frac{1}{11} \sum_{\zeta \in 0, 0.1, \cdots, 1} \max_{\hat{\zeta}:\hat{\zeta} > \zeta} PRE(\hat{\zeta})
\end{align*}
Here $\zeta$ represents recall threshold.
All metrics are calculated in BEV space.
\subsection{Training procedure}
In order to evaluate the algorithm in the training process,
we select 58 images from 289 training sets as validation sets.
In the process of algorithm analysis, in order to reduce computing resources and accelerate network training,
ResNet-18 is used as the basic network to verify the modules designed in this paper.
We use Adam algorithm to train the network with parameters shown in Table \ref{table_para}, and the learning rate is decaying linearly with training epoch.
 In the training process, the learning rate decreases linearly with time.
In the KITTI road experiment, we use ResNet-101\cite{7780459} and ResNet-50\cite{7780459} as the basic networks of the image and BEV branch respectively.
The two networks are trained separately first. During the training, symmetric transformation, translation transformation and scaling
 transformation  are employed for the data augmentation.
% For ResNet-101 branch, we use symmetric transformation, translation transformation, scaling transformation to expand the image.
% For ResNet-50 branch, random angle rotation, symmetric transformation and scale scaling transformation are used to extend the BEV image.

\begin{table}
  \centering
  \caption{Parameters of training}
  \scalebox{1.0}{
  \begin{tabular}{c c}
      % \hline
      \toprule[1.5pt]
      Parameter    & Value \\
      \hline
      Initial learning rate $\epsilon$  &  0.01 \\
      Momentum of learning rate $\eta$ &  0.9  \\
      Weight decay rate $\rho$   & 0.0001  \\
      Max training iteration $N$ & 30000 \\
      \bottomrule[1.5pt]
  \end{tabular}
  }
  \label{table_para}
\end{table}

\subsection{KITTI road experiment}
To compare with state-of-the-art road detection methods,
we estimate the BiFNet on the KITTI road benchmark \emph{validation} set. %\footnote{We have sent the application of the registration and the submission for the KITTI road benchmark. And we are waiting for the reply.}.
In this experiment, we use the ResNet-101 and ResNet-50 as the backbones for the camera image and
the BEV of the point cloud.
We use the 5-fold cross validation on the training set and obtain the mean of the results.
The overall results of BiFNet
and other state-of-the-art road detection methods are shown in Table \ref{tab_benchmark}.
LC-CRF\cite{gu2019road}, LidCamNet\cite{caltagirone2019lidar} and PLARD\cite{chen2019progressive} are the multi-sensor fusion-based methods.
LC-CRF fused the features from camera and perspective views of LiDAR while LidCamNet used the learnable weights to
add two different features. Since the proposed method has better scene adaptability, it achieves the better performance than the above two methods.
According to the results, our proposed method achieves state-of-the-art in UMM scenario and competitive results
in other scenarios.

Compared with PLARD, BiFNet achieves slightly better results according to Table \ref{tab_benchmark}. However, the results of PLARD shown
above are integrated from 3 different models, while BiFNet only needs single model.

\begin{table}
  \centering
  \caption{The comparsion results on the KITTI road validation set}
  \scalebox{0.95}{
    \begin{tabular}{l*{7} {c} c}
      % \hline
      \toprule[1.5pt]
      Method               &           & LC-CRF           & LidCamNet & PLARD  & BiFNet(Ours)  \\
      \hline
      \multirow{3}{*}{UM}  & MaxF     & 94.91\%          & 95.62\%             & \textbf{97.05\%}                   & 96.61\%  \\
                            \cline{2-6}
                             & AP      & 86.41\%          & 93.54\%             & 93.53\%                   & \textbf{94.31\%}     \\
      \hline
      \multirow{3}{*}{UMM} & MaxF     & 97.08\%          & 97.08\%              & 97.77\%                  & \textbf{97.88\%}   \\
                              \cline{2-6}
                             & AP      & 92.06\%          & 95.51\%              & 95.64\%                  & \textbf{95.82\%}   \\
      \hline
      \multirow{3}{*}{UU} & MaxF      & 94.01\%          & 94.54\%              & \textbf{95.95\% }        & 94.73\%   \\
                              \cline{2-6}
                             & AP      & 85.24\%          & 92.74\%              & \textbf{95.25\%}         & 93.31\%   \\
      % \hline
      \bottomrule[1.5pt]
    \end{tabular}
  }
  \label{tab_benchmark}
\end{table}

\subsection{Ablation study}

In order to verify the validity of different components of the proposed  BiFNet,
we conduct a set of ablation experiments on KITTI dataset.
The training set of KITTI is split into two parts. One is used to train the networks while
the other is for validation.

\begin{table}
  \centering
  \caption{Results of ablation Studies}
  \scalebox{1.0}{
  \begin{tabular}{c*{10} {c} c}
      % \hline
      \toprule[1.5pt]
      Image    & LiDAR BEV & DST     &  CBF     & MaxF &   AP  \\ %&  PRE  &  REC  &  FPR & FNR \\
      \hline
      $\surd$  &           &         &          & 93.01 & 94.41 \\ %& 92.22 & 93.81 & 8.34 & 6.19 \\
               & $\surd$   &         &          & 94.30 & 95.03 \\ %& 94.48 & 94.13 & 5.80 & 5.87 \\
      $\surd$  & $\surd$   & $\surd$ &          & 95.57 & 95.43 \\ %& \textbf{95.04} & 95.09 & \textbf{5.23} & 4.91 \\
      $\surd$  & $\surd$   & $\surd$ & $\surd$  & \textbf{96.21} & \textbf{96.08} \\%& 94.71 & \textbf{96.11} & 5.66 & \textbf{3.89} \\
      %$\surd$  & $\surd$   & $\surd$ & $\surd$ &  $\surd$    & 95.63 & 95.17 & 95.50 & 95.77 & 4.76 & 4.23 \\
      % \hline
      \bottomrule[1.5pt]
  \end{tabular}
  }
  \label{table_ablation}
\end{table}

\subsubsection{Baselines of image and LiDAR BEV}
In this part, we conduct single sensor-based road segmentation methods either with camera or LiDAR.
% These methods are implemented based on FCN-8s\cite{long2015fully}  framework and use ResNet as feature extraction networks.
% In order to verify the validity of the modules, we conduct a set of baseline experiments on KITTI dataset.
% The experimental results are shown in the table\ref{table_ablation}.
% Firstly, we remove the bidirectional fusion module and take ResNet as feature extraction networks
% to realize the image-based and LiDAR BEV-based road segmentation.
The results are shown in the first two lines of Table \ref{table_ablation}.
The first line shows the results based on images while the second line is LiDAR BEV-based.
We find that the networks with BEV input have better performance than those with image input
under the same network structure and training method.
There may be two reasons for these results. The first kind of illumination and road texture
have a greater impact on the network, resulting in poor network generalization performance.
The second is that perspective projection compresses distant objects, which results in poor segmentation accuracy.

\subsubsection{Effectiveness of DST}
Considering the above results, we add the DST module after ResNet backbone.
DST module can transform the features
from different space into the unique feature space. Then we use the element-wise addition
to fuse the information. The results are shown in third line of Table \ref{table_ablation}.
We find that the MaxF and AP of this method are much higher than the methods without DST.
Since the DST module gathers the information from the camera and LiDAR, the method with DST
is more robust for complex illumination environment.

\begin{figure*}
  \centering
  \includegraphics[width=0.98\textwidth]{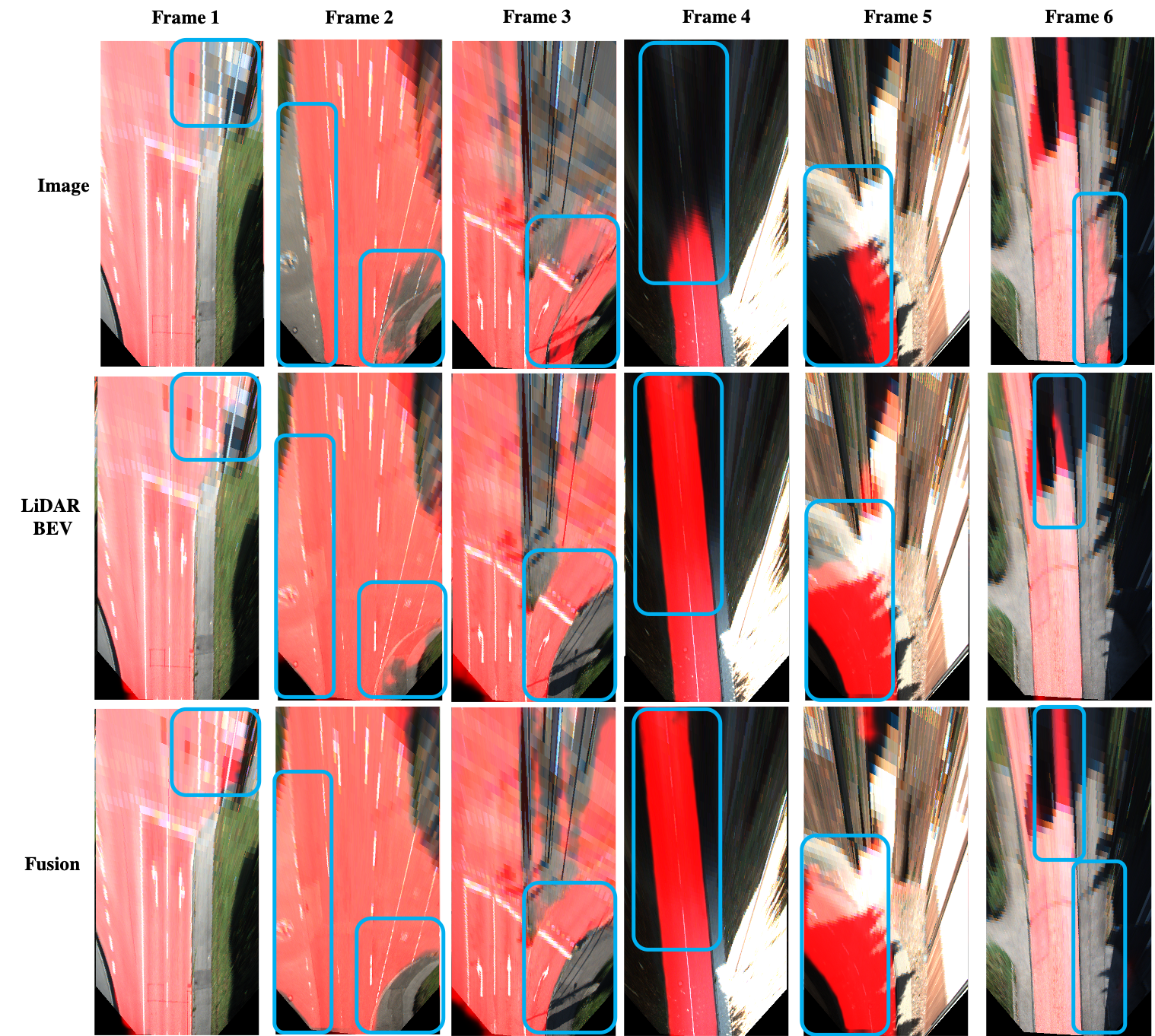}
  \caption{The results of the different methods on the image BEV.
  The rows are corresponding to the image-based method, LiDAR BEV-based method and the proposed fusion-based method.
  The columns are corresponding to the different scenes. The blue rectangle box indicates the region with different performances between the methods.}
\label{Fig_result}
\end{figure*}

\subsubsection{Effectiveness of CBF}
On the above basis, we add the CBF module after the DST module.
In last part, we employ the element-wise addition for fusing the gathered features from the image and LiDAR BEV by the DST module.
However, element-wise addition may confuse the features from different sensors.
For example, under dark environment, the image features are meaningless while the LiDAR features
are not affected. The direct addition will confuse the trained LiDAR features.
The CBF overcomes this problem and fuses the multi-sensors features based on
the corresponding expressiveness according to the context.
Compared with element-wise addition, CBF based method has a higher recall shown in the last line of Table \ref{table_ablation}.
The reason is that CBF fuses the information from multi-sensors based on the context features and has
higher utilization of information.
Compared with direct addition,
the CBF module improves the fusion efficiency and achieves better fusion results.

\subsubsection{Visualization of results}

In order to see more clearly the advantages of our proposed fusion algorithm compared with that based on a single sensor algorithm,
we visualize the results of different algorithms in different scenarios, as shown in Fig. \ref{Fig_result}.
Different rows represent different algorithms, and different columns represent different scenes.
As mentioned earlier, road segmentation based on camera space may result in poor remote segmentation results,
which is shown in Frame 1. It can be seen that in the LiDAR BEV-based method and fusion-based method,
the segmentation results of the distant intersection are significantly improved.
Frame 2 and Frame 3 show that the image-based method is more sensitive to the texture on the road,
but not to the height change on the road. The method based on LiDAR is more sensitive to the height change on the road surface.
Therefore, when there are steps at the fork, the method based on LiDAR is better than the method based on image,
and our method fully captures this advantage of LiDAR.
Frame 4 and Frame 5 show that light notes have a serious impact on image-based methods,
while LiDAR is not affected by them. In this case, our method tends to believe that features from LiDAR can segment the road.

\subsection{Comparison of space transformation}
Wange \emph{et al.}\cite{wang2018fusing} proposed sparse non-homogeneous pooling(SHPL) module to realize the fusion of image and LiDAR BEV.
Unlike DST in this paper, SHPL is based on the points of the point cloud to establish the relationship between image and BEV.
Because the points are sparse, this transformation is sparse. Only a small proportion of pixels in feature maps
have corresponding feature vectors.
In this paper, DST is a dense transformation, which realizes the transformation relationship of each pixel
of feature maps in different spaces.

In order to verify the validity of this dense transformation, we set up a comparison experiment with SHPL.
During the comparison, element-wise addition is used for features fusion.
% During the comparison, we use ResNet-18 as feature extraction networks and element-wise addition for features fusion.
The experimental results are shown in Table \ref{table_DST}.
From the results, we see that the two methods have the same effect on BEV branch.
For the perspective branch, DST outperforms SHPL significantly.
The reason is that DST is a dense spatial transformation method,
which can establish the mapping relationship between each pixel and the target feature at
different scales, and thus obtain dense transformation features.
In the experimental results, an interesting phenomenon is that with the sparse mapping,
SHPL achieves the good performance on BEV space.
One possible reason is that since the input LiDAR BEV is sparse,
the neural network has adapted to the sparsity in BEV space,
and has the ability to learn from the sparse input to segment the road,
but the perspective image is dense, and the network can not adapt to the sparsity from the point cloud.
This also proves the advantages of DST.

\begin{table}
  \centering
  \caption{Dense Space Transformation VS SHPL}
  \scalebox{1.0}{
      \begin{tabular}{l*{7} {c} c}
          % \hline
          \toprule[1.5pt]
          Method &  & MaxF & AP & PRE & REC  \\
          \hline
          \multirow{2}{*}{DST(Ours)}  & BEV           & \textbf{95.57} & \textbf{95.43} & \textbf{95.54} & 95.39  \\
                                \cline{2-6}
                                 & Perspective  & 94.14 & 94.20 & 92.85 & \textbf{95.46}      \\
          \hline
          \multirow{2}{*}{SHPL\cite{wang2018fusing}} & BEV           & 95.11 & 94.99  & 95.14  & 95.08   \\
                                  \cline{2-6}
                                 & Perspective  & 92.87  & 93.53 & 91.85 & 93.75    \\
          % \hline
          \bottomrule[1.5pt]
      \end{tabular}
  }
  \label{table_DST}
\end{table}

\subsection{CBF study}
In order to verify the validity and universality of our proposed CBF module, we reproduce LidCamNet\cite{caltagirone2019lidar} and PLARD\cite{chen2019progressive} methods,
and migrate the module to these methods.
The experimental results are shown in Table \ref{table_cbf}.
In the reproduction process, we use ResNet-18 as feature extraction network,
and the output features are used for fusion.
Unlike the experimental results in PLARD, our reproduced LidCamNet is slightly better than PLARD.
LidCamNet and PLARD both use the learnable weights for feature fusion.
The difference is that LidCamNet only learns the weights of LiDAR,
and each fusion connection has only one value.
The difference of PLARD is that it sets a learnable weight for each pixel position of the feature,
it is more adaptive than LidCamNet. However,
the fusion weights of these methods remain unchanged for all sample during the inference,
while the proposed method can adjust adaptively according to the content of samples.
Therefore, applying CBF module to the above two methods will gain performance to a certain extent.
% We find that CBF module has good gain effect when different forms of point cloud input are fused with images.

\begin{table}
  \centering
  \caption{The CBF influence on different networks}
  \scalebox{1.0}{
      \begin{tabular}{l*{10} {c} c}
          % \hline
          \toprule[1.5pt]
          Method                    & MaxF & AP & PRE & REC  \\
          \hline
          LidCamNet\cite{caltagirone2019lidar}  & 93.63 & 93.29 & 92.84 & 94.43 \\
          LidCamNet+CBF             & \textbf{94.02} & 93.97 & 93.18 & \textbf{94.88}  \\
          % LidCamNet+Multi-Scale AFF &       &    &     &     &     &     \\
          \hline
          PLARD\cite{chen2019progressive}  & 93.36 & \textbf{94.52} & 92.73 & 93.99  \\
          PLARD+CBF                 & 93.82 & 94.33 & \textbf{93.32 }& 94.33  \\
          % PLARD+Multi-Scale AFF     &       &    &     &     &     &     \\
          % \hline
          \bottomrule[1.5pt]
      \end{tabular}
  }
  \label{table_cbf}
\end{table}

\section{Conclusions}
In this paper, we propose a novel bidirectional fusion network named BiFNet which fuses the information of the camera image and the BEV of the point cloud.
Our innovative works for the network is to design the bidirectional fusion module which is composed of DST and CBF.
DST builds the dense spatial transformation relationship between the camera image and the BEV of the point cloud,
while CBF fuses the features from different sensors based on the various scene representation ability.
Moreover, CBF could be easily incorporated into deep neural networks and work for other forms of data.
The experimental tests on the KITTI road dataset also show that our method achieves a better result.

3D object detection plays a more and more important role in robot and automatic driving.
At present, the advanced 3D object detection methods\cite{ku2018joint}\cite{liang2018deep} are often based on BEV space.
This method faces great challenges in small object detection and object classification.
Multi-sensor fusion is the key to solve the problem.
Although BiFNet is used to solve the problem of road segmentation,
it can also be easily embedded into the existing 3D object detection algorithm.
Compared with the existing multi-sensor fusion detection methods,
our proposed method can be fused from the feature pixel level,
which is a potential way to improve the 3D detection algorithm.

% Furthermore, current advanced 3D object detection methods\cite{ku2018joint}\cite{liang2018deep} usually depend on the BEV space,
% BiFNet realizes the fusion of the BEV of LiDAR and the camera perspective space,
% while other methods fuses the features in the camera perspective space. Therefore, BiFNet is easier to extend to 3D detection methods.
\bibliographystyle{IEEEtran}
\bibliography{IEEEabrv, bfn}
\end{document}